\PassOptionsToPackage{bookmarks=false}{hyperref}
\documentclass[sigconf]{acmart}

\usepackage{booktabs} 
\usepackage{subcaption}
\graphicspath{{./figures/}}
\usepackage{amsmath}
\usepackage{amsfonts}       
\usepackage{bbm}

\usepackage{algorithmicx}
\usepackage{algorithm}
\usepackage{algpseudocode}

\usepackage[absolute]{textpos}

\newcommand{\bx}{\mathbf{x}}

\usepackage{multirow}
\usepackage{multicol}
\usepackage{booktabs}
\usepackage{nopageno} 
\usepackage{flushend}

\copyrightyear{2018} 
\acmYear{2018} 
\setcopyright{acmcopyright}
\acmConference[ICCAD '18]{IEEE/ACM INTERNATIONAL CONFERENCE ON COMPUTER-AIDED DESIGN}{November 5--8, 2018}{San Diego, CA, USA}
\acmBooktitle{IEEE/ACM INTERNATIONAL CONFERENCE ON COMPUTER-AIDED DESIGN (ICCAD '18), November 5--8, 2018, San Diego, CA, USA}
\acmPrice{15.00}
\acmDOI{10.1145/3240765.3243479}
\acmISBN{978-1-4503-5950-4/18/11}

\begin{document}

\begin{textblock}{10}(4.05,1.1)
\noindent{\footnotesize \normalfont This is the authors' final version. The authoritative version will appear in the proceedings of ICCAD 2018.}
\end{textblock}

\title{Hardware-Aware Machine Learning: Modeling and Optimization}
\subtitle{(\emph{Invited paper})}

\author{Diana Marculescu, Dimitrios Stamoulis, Ermao Cai\\
Department of ECE, Carnegie Mellon University, Pittsburgh, PA\\
Email: dianam@cmu.edu, dstamoul@andrew.cmu.edu, ermao@cmu.edu}
\renewcommand{\shortauthors}{D. Marculescu et al.}

\newcommand\blfootnote[1]{%
  \begingroup
  \renewcommand\thefootnote{}\footnote{#1}%
  \addtocounter{footnote}{-1}%
  \endgroup
}

\begin{abstract}
Recent breakthroughs in Machine Learning (ML) applications, and especially 
in Deep Learning (DL), have made DL models a key component in almost every modern 
computing system. The increased popularity of DL applications deployed on 
a wide-spectrum of platforms (from mobile devices to datacenters)
have resulted in a plethora of design challenges related to the constraints introduced by the hardware itself. ``What is the latency or energy cost for an inference made 
by a Deep Neural Network (DNN)?'' ``Is it possible to predict this latency or energy 
consumption before a model is even trained?'' ``If yes, how can machine 
learners take advantage of these models to design the hardware-optimal 
DNN for deployment?'' From lengthening battery life of mobile devices to 
reducing the runtime requirements of DL models executing in the cloud, 
the answers to these questions have drawn significant attention.

One cannot optimize what isn't properly modeled. Therefore, it is important to understand the hardware efficiency of DL models
during serving for making an inference, before even training the model. This 
key observation has motivated the use of predictive models to capture the 
hardware performance or energy efficiency of ML applications. Furthermore, ML practitioners are currently challenged with the task of designing the DNN model, 
\emph{i.e.}, of tuning the hyper-parameters of the DNN architecture, while 
optimizing for \emph{both} accuracy of the DL model and its hardware efficiency. 
Therefore, state-of-the-art methodologies have proposed \emph{hardware-aware} 
hyper-parameter optimization techniques. In this paper, we provide a comprehensive 
assessment of state-of-the-art work and selected results on the hardware-aware modeling and optimization for 
ML applications. We also highlight several open questions
that are poised to give rise to novel hardware-aware designs in the 
next few years, as DL applications continue to significantly impact associated hardware systems and platforms. 
\end{abstract}

\maketitle

\section{Introduction}

Recent advances in Deep Learning (DL) have enabled state-of-the-art results in 
numerous areas, such as text processing and computer vision. These breakthroughs in Deep Neural Networks (DNN) have been fueled by newly discovered ML model configurations ~\cite{bergstra2011algorithms} and advances in hardware platforms. 
However, the demand for better performance in real-world deployment results in increased complexity for both the hardware systems and the DL models.
As modern DNNs grow deeper and more complex, hardware constraints emerge as a key limiting factor. This ``hardware wall'' manifests its unintended, negative effects in several ways. 

First, the energy requirements of DNNs have emerged as a key 
impediment preventing their deployment on energy-constrained embedded and mobile
devices, such as Internet-of-Things (IoT) nodes and wearables. For instance, 
image classification with AlexNet \cite{krizhevsky2012imagenet} can drain the smartphone battery within an
hour ~\cite{yang2016designing}. This design challenge has resulted in 
a plethora of recent methodologies that focus on developing 
energy-efficient image classification solutions ~\cite{ILSVRC15}. However, 
identifying the most energy-efficient implementation on a given platform 
or the right DNN model-hardware platform pair can be challenging. Recent 
work shows that, while several DNN architectures can achieve a similar 
accuracy level~\cite{cai2017neuralpower}, the energy consumption differs drastically 
among these various iso-accuracy DNN configurations, many times by as much as 40$\times$. 

Second, there is an ever-increasing number of mobile applications that 
use on-device Machine Learning (ML) models employed directly into the smartphone, 
rather than relying on the cloud to provide the service. Thus, recent
work on the design of DNNs shows an ever-increasing interest in developing 
platform- and device-aware DL models~\cite{tan2018mnasnet, dong2018dpp}.
For instance, state-of-the-art hardware-aware methodologies from Google Brain
consider DL optimization and design techniques that optimize for both the 
accuracy of DNNs for image classification and the runtime of the model on 
Google Pixel smartphones~\cite{tan2018mnasnet}. 

Finally, edge-cloud communication constraints are an important 
design consideration for learning on the edge, \emph{e.g.},
for commercially important indoor localization applications based on 
low-cost sensing data (\emph{e.g.}, Radio-frequency identification RFID tags 
and WiFi signals)~\cite{rajendran2017distributed}. Hence, towards enabling hardware-aware ML applications, the aforementioned ``hardware wall'' gives rise to two main challenges:

\textbf{Challenge 1: Characterizing hardware performance of DL models}: 
The efficiency of DL models is determined by their hardware performance with respect
to metrics such as runtime or energy consumption, not only by their accuracy for a given learning task~\cite{tan2018mnasnet}. 
To this end, recent work explores modeling methodologies that aim to accurately model 
and predict the hardware efficiency of DNNs. In this paper, we review state-of-the-art
modeling tools, such as the ones employed in Eyeriss~\cite{isscc_2016_chen_eyeriss,yang2016designing}, or the Paleo~\cite{qi17paleo} and 
NeuralPower~\cite{cai2017neuralpower} frameworks.

\textbf{Challenge 2: Designing DL models under Hardware Constraints}: 
The \textbf{hyper-parameter optimization} of DL models, \emph{i.e.}, the
design of DL models via the tuning of hyper-parameters such as the number of layers
or the number of filters per layer, has emerged as an increasingly 
expensive process, dubbed by many researchers to be \emph{more of an art than a science}. 
Hyper-parameter optimization is a challenging design problem due to several
reasons. First, in commercially important Big Data applications (\emph{e.g.}, 
speech recognition), the construction of DL models involves many tunable hyper-parameters~\cite{shahriari2016taking} and the training of each configuration
takes days, if not weeks, to fully train. More importantly, 
the ability of a human expert to identify the best performing DL 
model could be hampered significantly if we are to consider hardware constraints 
and design considerations imposed by the underlying hardware~\cite{stamoulis2017hyperpower}.
To this end, there is an ever-increasing interest in works that co-optimize for 
both the hardware efficiency and the accuracy of the DL model. 
In this paper, we investigate the main tools for employing hardware-aware 
hyper-parameter optimization, such as methodologies based on hardware-aware 
Bayesian optimization~\cite{stamoulis2017hyperpower, stamoulis2018designing}, multi-level co-optimization~\cite{reagen2016minerva} and 
Neural Architecture Search (NAS)~\cite{tan2018mnasnet, dong2018dpp}.

\section{Related work}

Reducing the complexity of the ML models has long been a concern for machine learning practitioners. Hence, while this paper focuses on 
hardware-aware modeling and optimization methodologies, there are orthogonal 
techniques that have been previously explored as means to reduce the 
complexity of DL models. In this section, we briefly review these approaches
as they relate to hardware-aware modeling and optimization, and we highlight
their limitations. 

\textbf{Pruning}: Prior art has investigated pruning-based methods that
aim at reducing the ML model complexity by gradually removing weights
of the model and by retraining it~\cite{han2015learning, dai2017nest, yang2016designing}. 
The key insight behind this type of work, \emph{e.g.}, deep compression~\cite{han2015deep}, 
is the fact that modern DL models exhibit redundancy to their features and a small degradation of accuracy can be traded off for a significant reduction in the computational cost. This insight has resulted in a significant body of work that 
incorporates regularizing terms directly in the objective function used for the training
of DL model training. These loss regularizers, \emph{e.g.}, L1 norm terms, ``zero out'' 
several connections, thus reducing the overall model complexity directly at training. 
For instance, in~\cite{gordon2017morphnet}, the authors propose MorhpNet, a
resource-constrained regularization that outperforms prior methodologies in terms of accuracy given a maximum FLOP constraint. Nonetheless, these methods use the number of floating points operations or the number of model parameters as a proxy for the model complexity, without explicitly accounting for hardware-based metrics such as energy or power consumption. 

\textbf{Quantization}: Beyond pruning-based methodologies 
that reduce the DL model complexity, 
other works reduce directly their computational complexity via 
quantization~\cite{courbariaux2016binarized}. The key insight is that, 
while the number of FLOPs remains the same, the hardware execution cost
is reduced by reducing the bit-width per multiply-accumulate (MAC) operation.
In turn, the reduced cost per operation results in overall hardware efficiency, compounded by lower storage requirements. 
Several methodologies have explored this trade-off by considering different
bit-width values~\cite{gupta2015deep, ding2017lightnn, ding2018quantized, jacob2017quantization}. Nevertheless,
the effectiveness of quantization methods relies heavily on the performance of 
the pre-quantized (seed) network~\cite{gordon2017morphnet}, without providing
any insight on how the sizing of different DL hyper-parameters affects the
overall hardware efficiency and accuracy of the DL model.

\textbf{Discussion}: While pruning- and quantization-based methods
have been shown to significantly reduce the computational complexity 
of ML models, recent work has highlighted some key limitations. 
First, these methodologies
focus on reducing the storage required by model weights or the number of FLOPs. However, 
Yang \emph{et al.} show that FLOP reduction does not yield the optimal DNN design with respect to power or energy 
consumption~\cite{yang2016designing} largely because the parameter-heavy layers
of a DL model are not necessarily the most energy-consuming. Hence, 
pruning and quantization methodologies rely on formulations that are hardware-unaware
and they do not necessarily result in Pareto-optimal configurations
in terms of hardware efficiency, even though the targeted FLOP constraint
is satisfied. In contrast, hardware-aware modeling methodologies 
can pave the way to directly account for energy consumption or runtime, as discussed in the next section. 

Second, pruning methodologies traverse the design space only locally
around the original pre-trained DL model used as a seed. Consequently, 
if the configuration of the DL model is viewed as a hyper-parameter optimization
problem to be globally solved, pruning methodologies can only reach locally optimal 
solutions, whose effectiveness is bound by the quality of the seed design. 
Nevertheless, Tan \emph{et al.} show that that hardware-constrained models such 
as MobileNets~\cite{howard2017mobilenets} are not Pareto-optimal with respect to 
accuracy for a given level of hardware complexity. Instead, state-of-the-art 
results in platform-aware hyper-parameter optimization~\cite{tan2018mnasnet} 
yield DL design closer to the Pareto front by globally solving the hardware-aware
optimization problem. In this paper, we therefore focus on hardware-aware 
optimization techniques based on Bayesian optimization and Neural Architecture Search (NAS).

\section{Hardware-aware Runtime and Energy Modeling for DL}

\subsection{DL model-hardware platform interplay}
As mentioned before, effective optimization requires efficient and reliable models. It is not surprising, therefore, that the first efforts in hardware models for DL have been developed in the context of hardware efficient DNNs. In~\cite{yang2016designing}, Yang \emph{et al.} have shown that energy-based 
pruning achieves better energy-efficiency compared to its FLOPs-based counterpart.
During each iteration of pruning-finetuning, Yang \emph{et al.} use an energy
consumption model to decide on which DNN layers to prune next. To this end, 
the authors develop a predictive model based on measurements on their hardware
accelerator, namely Eyeriss~\cite{isscc_2016_chen_eyeriss}. Their formulation
models energy consumption as a weighted function of the number of accessed to the
different levels of the memory hierarchy, wherein each level is profiled based 
on the energy consumption per operation or per memory access pattern (\emph{e.g.}, 
overhead to access the register file, the global buffer, the main memory, \emph{etc.}). 

\begin{figure*}[ht!] 
	\centering
	\small
	\includegraphics[width=0.7\linewidth]{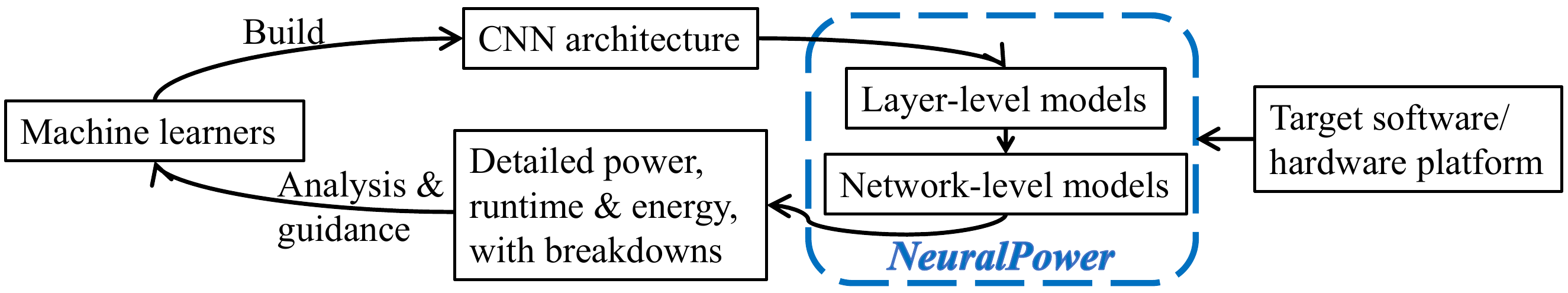}	
	\vspace{-10pt}
	\caption{ Overview of NeuralPower \cite{cai2017neuralpower}.}
	\vspace{-6pt}
	\label{fig:neural_power}
\end{figure*}

Specifically, they calculate the energy consumption for each layer in two parts, computation energy consumption ($E_{comp}$), and data movement energy consumption ($E_{data}$). $E_{comp}$ is calculated by multiplying the number of MACs in the layer by the energy consumption for running each MAC operations in the computation core. $E_{data}$ is the total summation of energy consumption of all the memory access for each level of energy. In addition, they account for the impact of data sparsity and bitwidth reduction on energy consumption. This is based on their assumption that a MAC and its memory access can be skipped completely when the input activation or weight is zero. The impact of bitwidth is also calculated by scaling the energy consumption of corresponding hardware units.

This work provides a good model to evaluate the energy cost for running DNNs on the ASICs. However, transferring the model or methodology to other platforms is not trivial as it may require additional model selection and training.

In a more recent development, the Paleo framework proposed by Qi \emph{et al.}~\cite{qi17paleo} proposes a methodology for modeling of hardware metrics, \emph{i.e.}, 
runtime of DNNs, without resorting to FLOPs as a proxy for hardware efficiency. As opposed to Eyeriss, Paleo has been developed with general purpose platforms (GPUs) in mind. In their approach, the authors present an analytical method 
to determine the runtime of DNNs executing on various platforms. The framework
has been shown to flexibly scale across different GPU platforms and DNN types. 

Specifically, Paleo divides the runtime for each layer into two parts, computation time and communication time. For a layer $u$, and the operation $f$ on a device $d$, the total execution time $T(u)$ can be expressed as:
\begin{equation}
T(u) = \mathcal{R}(Pa(u)) + \mathcal{C}(f,d) + \mathcal{W}(f,d)
\end{equation}
where $\mathcal{R}(Pa(u))$ is the time to fetch the input produced by its parent layers, $\mathcal{W}(f,d)$ is the time to write the outputs to the local memory, and $\mathcal{C}(f,d)$ is the total computation time with $f$. Assuming that we are interested in metrics for a device $d$ under average speed conditions, the computation time can be determined by: $\mathcal{C}(f,d) = \text{FLOPs}(f) / speed(d)$. Following a similar assumption on IO bandwidth, $\mathcal{R}$ and $\mathcal{W}$ can be calculated as the amount of the reading/writing data divided by the average IO bandwidth. In the case of multiple workers, the model still holds by replacing the average IO bandwidth with the communication bandwidth between two devices.

Since the peak FLOPs and peak bandwidth provided by the manufacturers are usually different than the actual speed of the devices running specific DNNs, the authors introduce the concept of \emph{platform percent of peak} (PPP) to capture the average relative inefficiency of the platform compared to peak performance. However, for different devices, the average computation and communication speeds can vary from one to another. Therefore, one needs to evaluate those metrics with many different tests to obtain reliable values. In addition, Paleo does not consider models for predicting power or energy consumption.

\subsection{The NeuralPower framework}
To provide adaptive and reliable prediction of runtime, power, and energy for DNNs simultaneously, Cai \emph{et al.}~\cite{cai2017neuralpower} have introduced NeuralPower, 
a layer-wise predictive framework based on sparse polynomial regression for determining the serving energy consumption of convolutional neural networks (CNNs) deployed
on GPU platforms. Given the architecture of a CNN, NeuralPower provides an accurate
prediction and breakdown for power and runtime across all layers in the whole network.
In their framework, the authors also provide a network-level model for the energy consumption
of state-of-the-art CNNs. 

Specifically, NeuralPower introduces a flexible and comprehensive way to build the runtime, power, and energy models for various DNNs on a variety of platforms. The modeling can be divided into two parts, as shown in Figure \ref{fig:neural_power}. 

The layer-level approach provides accurate models for running a specific layer on each of the considered platform. Instead of using proxies to determine how fast a layer runs as in Paleo, NeuralPower learns the models by actually running the DL models on the target platform. For example, the runtime $\hat{T}$ of a layer can be expressed as: 
\begin{align}
  \label{eq:polynomial_runtime}
  \hat{T}(\bx_T)  =  & \sum _{j}  c_j \cdot  \prod_{i = 1}^{D_T} {x}_i^{q_{ij}} + \sum_s c^\prime_s \mathcal{F}_s(\bx_T)\\
\text{where  } & \bx_T \in \mathbb{R}^{D_T}; \ q_{ij} \in \mathbb{N}; \  \forall j, \ \sum_{i = 1}^{D_T} q_{ij} \leq K_T. \nonumber
\end{align}

The model in Equation \ref{eq:polynomial_runtime} consists of two parts. The first part is the regular degree-$K_T$ polynomial terms which are a function of the features in the input vector $\bx_T \in \mathbb{R}^{D_T}$. $x_i$ is the $i$-th component of $\bx_T$. $q_{ij}$ is the exponent for $x_i$ in the $j$-th polynomial term, and $c_j$ is the coefficient to learn. This feature vector of dimension $D_T$ includes layer configuration hyper-parameters, such as the batch size, the input size, and the output size. For different types of layers, the dimension $D_T$ is expected to vary. For convolutional layers, for example, the input vector includes the kernel shape, the stride size, and the padding size, whereas such features are not relevant to the formulation/configuration of a fully-connected layer. The second part is comprised of special polynomial terms $\mathcal{F}$, which represent physical operations related to each layer (\emph{e.g.}, the total number of memory accesses and the total FLOPs). The number of the special terms differs from one layer type to another. Finally, $c^\prime_s$ is the coefficient of the $s$-th special term to learn.

Similarly, a power model can be constructed and the total energy consumption can be calculated by the product of power consumption and runtime. The models are trained on the real data obtained from real GPU platforms running state-of-the-art DNNs. 

One can compare NeuralPower against Paleo with respect to runtime prediction only, as the latter does not address energy or power modeling. Table \ref{tab:perf_model} shows the modeling results for each layer. As it can be seen, NeuralPower outperforms Paleo in terms of model accuracy for the most widely used types of layers in CNNs.
\begin{table}[ht]	
	\centering
	\caption{Comparison of runtime models for common CNN layers between NeuralPower \cite{cai2017neuralpower} and Paleo \cite{qi17paleo}.} 
		\vspace{-5pt}
	\label{tab:perf_model}
		\small
		\begin{tabular}
			{l|ccc|cc} \toprule
 			\multirow{2}{*}{Layer} & \multicolumn{3}{c}{NeuralPower \cite{cai2017neuralpower}} & \multicolumn{2}{|c}{Paleo~\cite{qi17paleo}} \\  \cline{2-4}\cline{5-6}
            & Model size & RMSPE & RMSE (ms) & RMSPE & RMSE  (ms)\\ \hline
			CONV & 60 & 39.97\%	& 1.019  & 58.29\% & 4.304\\ 
			FC & 17 & 41.92\% & 0.7474 & 73.76\% & 0.8265\\ 
			Pool & 31 & 11.41\% & 0.0686  & 79.91\% &1.763 \\ \bottomrule
		\end{tabular}	 
\end{table}

For the network level model, NeuralPower uses the summation of the layer level results to achieve total runtime and total energy. The average power can be calculated by dividing total energy by total runtime. Table \ref{tab:whole_model_runtime} shows the runtime predictions at network level for NeuralPower and Paleo. From Table \ref{tab:perf_model} and Table \ref{tab:whole_model_runtime}, we can see NeuralPower generally achieves better
accuracy in both layer-level and network-level runtime prediction.
\begin{table}[ht]	
	\centering
	\caption{Comparison of runtime models for common CNNs between NeuralPower \cite{cai2017neuralpower} and Paleo \cite{qi17paleo}.}
		\vspace{-5pt}
	\label{tab:whole_model_runtime}
		\small
		\begin{tabular}
			{c|c|c||c} \toprule
			CNN  & Paleo\cite{qi17paleo} & NeuralPower\cite{cai2017neuralpower}    & Actual runtime  \\ \hline
			VGG-16 & 345.83 & 373.82 &   368.42 \\ 
			AlexNet & 33.16  & 43.41 & 39.02 \\ 
			NIN & 45.68  & 62.62 &  50.66 \\ 
			Overfeat & 114.71  & 195.21 &  197.99 \\ 
			CIFAR10-6conv & 28.75  & 51.13 & 50.09 \\ \bottomrule
		\end{tabular}	 
\end{table}

Lastly and most importantly, NeuralPower is currently the state-of-the-art method in
terms of prediction error when capturing the runtime, power, and energy consumption 
for various CNNs, with mean squared error less than $5\%$. Moreover, 
NeuralPower achieves a $70\%$ accuracy improvement compared the previously best
model, \emph{i.e.}, Paleo. Using learning-based polynomial regression models, NeuralPower
can be flexibly employed for prediction across different DL software tools (\emph{e.g.},
Caffe~\cite{jia2014caffe} or TensorFlow~\cite{abadi2016tensorflow}) and GPU platforms
(\emph{i.e.}, both low-power Nvidia Tegra boards and workstation Nvidia Titan X GPUs).

\section{Hardware-aware Optimization for DL}

\subsection{Model-based optimization}
\begin{figure}
  \centering
  \includegraphics[width=0.90\columnwidth]{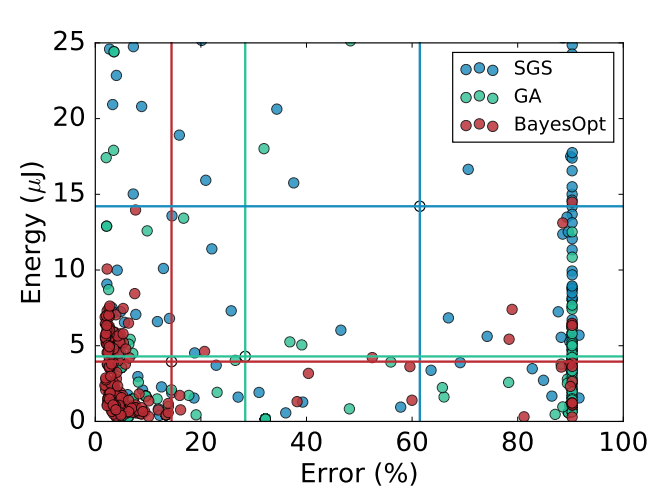}
  \vspace{-10pt}
  \caption{Model-based optimization (denoted as BayesOpt) outperforms
alternative methods such as random search and genetic algorithms in terms of 
finding more accurate, energy-efficient designs closer to the Pareto front~\cite{reagen2017case}.}
  \vspace{-10pt}
  \label{fig:lobato}
\end{figure}
\textbf{Background}: Even without 
hardware-related design objectives, the hyper-parameter optimization
of DL models, \emph{i.e.}, deciding on tunable hyper-parameters such as the number
of features per layer or the number of layers, is a challenging design task. 
Early attempts towards efficiently solving this problem use sequential model-based
optimization (SMBO) when the objective function (\emph{i.e.}, test error of each 
candidate DNN configuration) has no simple closed form and its evaluations
are costly. SMBO methodologies use a surrogate (cheaper to evaluate) probabilistic 
model to approximate the objective function.

Different formulations have been used for SMBO probabilistic models, such 
as Gaussian Processes (GP)~\cite{shahriari2016taking} or tree-structured Parzen estimators  
(TPE)~\cite{bergstra2011algorithms}. Regardless of the choice of the SMBO formulation,
intuitively the probabilistic model encapsulates the
belief about the shape of functions that are more likely to fit the data observed so far, 
providing us with a cheap approximation for the mean and the uncertainty of the objective
function. 

Each SBMO algorithm has three main steps: (i) \emph{maximization of acquisition function}: to select the point (\emph{i.e.}, next candidate DNN configuration) at which the objective will be evaluated next, SMBO methods use the so-called acquisition function.
Intuitively, the acquisition function provides the direction in the design
space toward which there is an expectation of improvement of the objective.
By evaluating cheaply the acquisition function at different candidate points (\emph{i.e.},
different DNN configurations), SMBO selects the point with maximum acquisition function value.
(ii) \emph{evaluation of the objective}: the currently selected DNN design is created
and trained to completion to acquire the validation error. (iii) \emph{probabilistic 
model update}: based on the value of the objective at the currently considered DNN design, 
the probabilistic model is updated. 

\textbf{Hardware-aware SMBO}: If Gaussian processes are used to formulate
the probabilistic model, the resulting special case
SMBO constitutes a powerful approach, namely Bayesian optimization~\cite{shahriari2016taking}.
Prior art has proposed general frameworks for employing Bayesian optimization 
with multiple objective and constraint terms~\cite{hernandez2016general}.
To this end, a natural extension is to use these formulations to solve the 
hyper-parameter optimization of DNNs under hardware constraints. 

Hern{\'a}ndez-Lobato \emph{et al.} has successfully used Bayesian optimization
to design DNNs under runtime constraints~\cite{hernandez2016general}. More interestingly,
the authors have investigated the co-design of hardware accelerators and DNNs 
in~\cite{hernandez2016designing, reagen2017case}, where the energy consumption
and the accuracy of a DNN correspond to objective terms that depend 
on both DNN design choices (\emph{e.g.}, number of features per layer or 
number of layers) and hardware-related architectural decisions (\emph{e.g.}, 
bit width and memory size).

\begin{figure*}
  \centering
  \includegraphics[width=7.1in, height=1.3in]{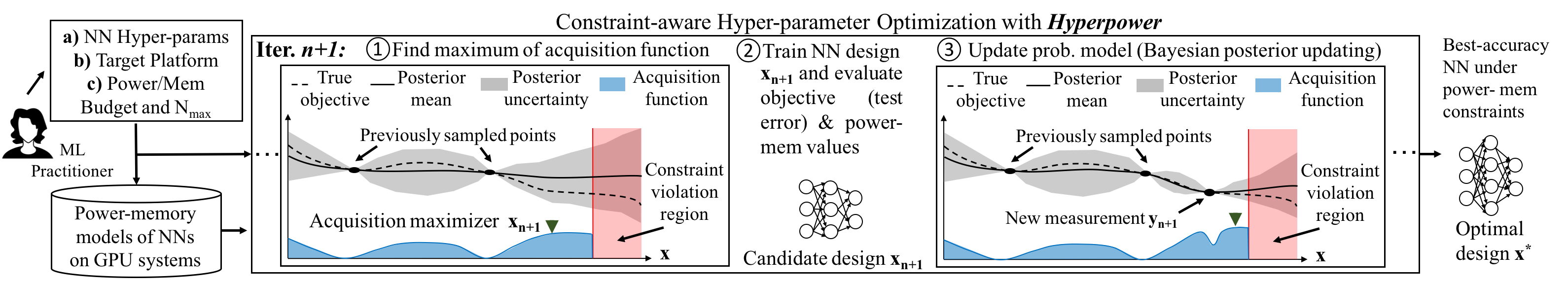}
  \vspace{-10pt}
  \caption{Overview of HyperPower flow and illustration of the Bayesian optimization
  procedure during each iteration \cite{stamoulis2017hyperpower}.}
  \vspace{0pt}
  \label{fig:hyperpower}
\end{figure*}

In Figure~\ref{fig:lobato}, we can observe that the SMBO-based method
(denoted as BayesOpt) outperforms alternative methods such as random search and genetic algorithms in terms of 
finding more accurate, energy-efficient designs closer to the Pareto front.
This is to be expected, since the model-based approach provides a hardware-aware
model to guide the design space exploration faster towards Pareto-optimal designs.

\subsection{Multi-layer co-optimization}

From a different perspective, Minerva \cite{reagen2016minerva} presents an automated co-optimization method including multiple layers in the system stack: algorithm, architecture and circuit levels. In that approach, five stages are used: 1. Training space exploration; 2. Microarchitecture design space; 3. Data type quantization; 4. Selective operation pruning; 5. SRAM fault mitigation. Based on these five steps, the DNNs can be trained, pruned and deployed to Minerva with low power consumption, without sacrificing their model accuracy.  


In general, the savings from optimization at the software, architecture, and circuit level, can be combined together. According to tests on five different datasets, including MNIST, Forest, Reuters, WebKB, and 20NG, Minerva achieves an average power saving of $8.1\times$ compared with a baseline accelerator. The authors conclude that fine-grain, heterogeneous datatype optimization, aggressive prediction and pruning of small activity values, and active hardware fault detection coupled with domain-aware error mitigation are the main contributors for the savings. Given its low footprint, Minerva can be applied to IoT and mobile devices, but it doesn't rely on a comprehensive optimization framework that can be extended to generic platforms. We present next one such approach.

\subsection{HyperPower framework}

A special case of Bayesian optimization is one where constraints can be expressed and known \emph{a priori}; these formulations enable models that can directly capture candidate configurations as valid or invalid~\cite{gelbart2014bayesian}. In HyperPower~\cite{stamoulis2017hyperpower}, 
Stamoulis \emph{et al.} investigate the key insight that predictive models 
(as discussed in the previous section) can provide an \emph{a priori} knowledge 
on whether the energy consumption of any DNN configuration violates the
energy budget or not. In other words, predictive models can be used
in the context of Bayesian optimization and can be formulated as 
\emph{a priori} known constraints. 

More specifically, HyperPower introduces a hardware-aware acquisition function
that returns zero for all constraint-violating candidate designs, hence effectively 
guiding the design space exploration toward hardware constraint-satisfying configurations
with optimal DNN accuracy. By accounting for hardware constraints directly in the
SMBO formulation, HyperPower reaches the near-optimal region $3.5 \times$ 
faster compared to hardware-unaware Bayesian optimization.
The proposed acquisition function uses predictive models for 
power consumption and memory utilization, thus showing the interplay of both
modeling and optimization towards enabling hardware-aware DL.

The general structure of HyperPower is shown in Figure \ref{fig:hyperpower}. Bayesian optimization is a sequential model-based approach that approximates the objective function
with a surrogate (cheaper to evaluate) probabilistic model $\mathcal{M}$, based on Gaussian processes (GP).
The GP model is a probability distribution over the possible functions of $f(\textbf{x})$, and it
approximates the objective at each iteration $n+1$ based on
data $\textbf{X}:= {x_i \in \mathcal{X}}_{i=1}^{n}$ queried so far.
HyperPower assumes that the values $\textbf{f} := f_{1:n}$
of the objective function at points $\textbf{X}$ are jointly Gaussian
with mean $\textbf{m}$ and covariance $\textbf{K}$, \emph{i.e.},
$\textbf{f} ~| ~\textbf{X} \sim  \mathcal{N}(\textbf{m}, \textbf{K})$.
Since the observations $\textbf{f}$ are noisy
with additive noise $\epsilon \sim  \mathcal{N} (0,\sigma^2)$, the GP model can be written as 
$\textbf{y} ~|~ \textbf{f}, \sigma^2 \sim \mathcal{N} (\textbf{f} , \sigma^2 \textbf{I})$.
At each point $\textbf{x} $, GP provides a cheap approximation
for the mean and the uncertainty of the objective, written
as $p_\mathcal{M} (y|\textbf{x})$ and illustrated in Figure~\ref{fig:hyperpower} with the black curve and
the grey shaded areas.

Each iteration $n+1$ of a Bayesian optimization algorithm consists of three key steps:

\textbf{Maximization of acquisition function}:
one first needs to select the point $\textbf{x}_{n+1}$
(\emph{i.e.}, next candidate NN configuration) at which the objective (\emph{i.e.}, the test error of the candidate NN)
will be evaluated next. This task of guiding the search relies on the
so-called acquisition function $\alpha(\textbf{x})$. 
A popular choice for the acquisition function is the Expectation Improvement (EI) criterion,
which computes the probability that the objective function $f$ will exceed (negatively) some threshold $y^+$,
\emph{i.e.}, $EI(\textbf{x}) =  \int_{-\infty}^{\infty} \max\{y^+ - y, 0\} \cdot p_\mathcal{M} (y|\textbf{x}) ~ dy$.
Intuitively, $\alpha(\textbf{x})$ provides a measure of the direction toward
which there is an expectation of improvement of the objective function.

The acquisition function is evaluated at different candidate points $\textbf{x}$, yielding
high values at points where the GP's uncertainty is high (\emph{i.e.}, favoring exploration),
and where the GP predicts a high objective (\emph{i.e.}, favoring exploitation)~\cite{shahriari2016taking};
this is qualitatively illustrated in Figure~\ref{fig:hyperpower} (blue curve).
HyperPower selects the maximizer of $\alpha(\textbf{x})$ as the point $\textbf{x}_{n+1}$ to evaluate next
(green triangle in Figure~\ref{fig:hyperpower}).
To enable power- and memory-aware Bayesian optimization, \emph{HyperPower} incorporates
hardware-awareness directly into the acquisition function.

\textbf{Evaluation of the objective}: Once
the current candidate NN design $\textbf{x}_{n+1}$ has been selected, the NN is generated
and trained to completion to acquire the test error. This is the most expensive step.
Hence, to enable efficient Bayesian optimization, HyperPower focuses on detecting when this step can be bypassed.

\textbf{Probabilistic model
update}: As the new objective value $y_{n+1}$ becomes available at the end
of iteration $n+1$, the probabilistic model
$p_\mathcal{M} (y)$ is refined via Bayesian posterior updating
(the posterior mean $\textbf{m}_{n+1} (\textbf{x})$ and covariance covariance $\textbf{K}_{n+1}$
can be analytically derived).
This step is quantitatively illustrated in Figure~\ref{fig:hyperpower} with the black curve and
the grey shaded areas. Attention can be paid to how the updated model has reduced uncertainty around the previous samples and
newly observed point. 
For an overview of GP models 
the reader is referred to~\cite{shahriari2016taking}. 

To enable \emph{a priori} power and memory constraint evaluations that are
decoupled from the expensive objective evaluation,
HyperPower models power and memory consumption of a network as a function of
the $J$ discrete (structural) hyper-parameters $\textbf{z} \in \mathbb{Z}_{+}^J$ (subset of
$\textbf{x} \in \mathcal{X}$); it trains on the structural hyper-parameters
$\textbf{z}$ 
that affect the NN's power and memory (\emph{e.g.}, number of hidden units),
since parameters such as learning rate have negligible impact.

To this end, HyperPower employs offline random sampling by generating
different configurations based on the ranges of the considered
hyper-parameters $\textbf{z}$.
Since the Bayesian optimization corresponds to function evaluations with respect to the
test error\cite{snoek2012practical}, for each candidate design $\textbf{z}_l$
HyperPower measures the hardware platform's power $P_l$ and memory $M_l$ values during inference and
not during the NN's training. Given the $L$ profiled data points $\{(\textbf{z}_l,
P_l, M_l)\}_{l=1}^L$, the following models that are linear with respect to both
the input vector $\textbf{z} \in \mathbb{Z}_{+}^J$ and model weights $\textbf{w},\textbf{m}
\in \mathbb{R}^J$ are trained, \emph{i.e.}:
\vspace{-7pt}
\begin{equation}
\label{eq:pow}
\textbf{Power model}:~~\mathcal{P}(\textbf{z}) = \sum_{j=1}^{J} w_j \cdot z_j
\end{equation}
\vspace{-7pt}
\begin{equation}
\label{eq:mem}
\textbf{Memory model}:~~\mathcal{M}(\textbf{z}) = \sum_{j=1}^{J} m_j \cdot z_j
\end{equation}
HyperPower trains the models above by employing a 10-fold cross validation on the dataset $\{(\textbf{z}_l,
P_l, M_l)\}_{l=1}^L$. While the authors experimented with nonlinear regression formulations which can be
plugged-in to the models (\emph{e.g.}, see recent work~\cite{cai2017neuralpower}),
these linear functions provide sufficient accuracy. More importantly,
HyperPower selects the linear form since it allows for the efficient evaluation of
the power and memory predictions within the acquisition function (next subsection), computed
on each sampled grid point of the hyper-parameter space. 

\textbf{\emph{HW-IECI}}: In the context of hardware-constraint optimization, EI allows to
directly incorporate the \emph{a priori} constraint
information in a representative way. Inspired by constraint-aware 
heuristics~\cite{gelbart2014bayesian}~\cite{gramacy2010optimization},
the authors propose a power and memory constraint-aware acquisition function:
\begin{equation}
\label{eq:con_ei}
\begin{split}
a(\textbf{x}) =  &\int_{-\infty}^{\infty} \max\{y^+ - y, 0\} \cdot p_\mathcal{M} (y|\textbf{x}) \cdot \\
&  \mathbb{I}[\mathcal{P}(\textbf{z}) \leq \text{PB}] \cdot \mathbb{I}[\mathcal{M}(\textbf{z}) \leq \text{MB}] ~ dy
\end{split}
\end{equation}
where $\textbf{z}$ are the structural hyper-parameters, 
$p_\mathcal{M} (y|\textbf{x})$ is the predictive marginal density
of the objective function at $\textbf{x}$ based on surrogate model $M$.
$\mathbb{I}[\mathcal{P}(\textbf{z}) \leq \text{PB}]$ and $\mathbb{I}[\mathcal{M}(\textbf{z}) \leq
\text{MB}]$ are the indicator functions, which are equal to $1$ if
the power budget PB and the memory budget MB are respectively satisfied.
Typically, the threshold $y^+$ is adaptively set to the best value
$y^+ = \max_{i=1:n} y_i$ over previous observations~\cite{shahriari2016taking}\cite{gelbart2014bayesian}.

HyperPower captures the fact that improvement should not
be possible in regions where the constraints are violated.
Inspired by the integrated expected conditional improvement
(IECI)~\cite{gramacy2010optimization} formulation, the authors refer to this
proposed methodology as \emph{HW-IECI}. Uncertainty can be also encapsulated by replacing the indicator functions with
probabilistic Gaussian models as in~\cite{gramacy2010optimization},
whose implementation is already supported by
the used tool~\cite{snoek2012practical}.

\begin{figure}
  \centering
  \includegraphics[width=0.90\columnwidth]{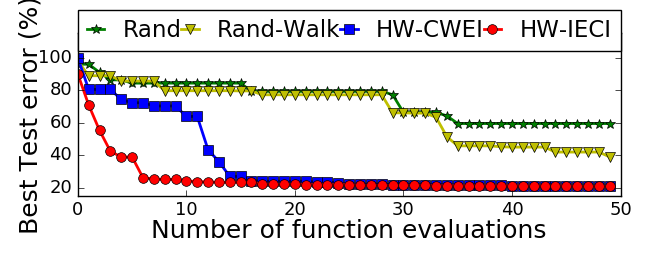}
  \vspace{-10pt}
  \caption{Assessment of HyperPower (denoted as HW-IECI) against
  hardware-unaware methods. Best observed test error against the number of
function evaluations \cite{stamoulis2017hyperpower}.}
  \vspace{-10pt}
  \label{fig:hyperpower-results}
\end{figure}

NeuralPower exploits
the use of trained predictive models based on the power and
memory consumption of DNNs, allowing the HyperPower framework
to navigate the design space in a constraint ``complying''
manner. As shown in Figure \ref{fig:hyperpower-results}, the authors observe that the proposed HyperPower 
methodology (denoted as HW-IECI) converges to the high-performing, constraint-satisfying 
region faster compared to the hardware unaware one.

A key advantage of model-based optimization methodologies is that, given 
their black-box optimization nature, they can be extended to various type
of hardware constraints and design considerations. For instance, 
Stamoulis \emph{et al.} \cite{stamoulis2018designing} show that Bayesian 
optimization can be used to efficiently design adaptive DNNs under
edge-node communication constraints. In particular, hardware-aware adaptive
DNNs achieve $6.1\times$ more energy efficient designs
than state-of-the-art for same accuracy.

\subsection{Platform-aware NAS}

Traditional SMBO-based black-box optimization methods fit a probabilistic model
over the entire design, \emph{i.e.}, the hyper-parameters of all layers can be
simultaneously changed. Hence, while Bayesian optimization methodologies are extremely 
efficient for design spaces with up to 20 hyper-parameters, they could suffer
from high dimensionality in larger DL models, \emph{e.g}, parameter-heavy 
Recurrent Neural Networks (RNNs) used in speech recognition applications. 

To address this challenge, recent breakthroughs in hyper-parameter optimization focus on 
the following insight: DL models designed by human experts exhibit pattern repetition, such as the residual cell in ResNet-like configurations~\cite{he2016deep}.
To this end, recent work focuses on the hyper-parameter optimization of a repetitive 
motif, namely cell, which is identified on a simpler dataset. After the optimal cell 
is identified, it is repeated several times to construct larger, more complex DL models
that achieve state-of-the-art performance on larger datasets and more complex 
learning tasks. Intuitively, this transferability property of the optimal cell 
allows the hyper-parameter optimization probabilistic model to focus on a smaller structured
design spaces with fewer hyper-parameters. Recent work uses techniques such as Reinforcement 
learning (RL) or Evolutionary algorithms (EA) to guide the exploration towards the optimal 
cell. The body of work based on this insight is called Neural Architecture
Search (NAS). Nevertheless, initial NAS-based methods focused explicitly on
optimizing accuracy, without considering hardware-related metrics. 

Motivated by this observation, the recently published NAS frameworks
MnasNet~\cite{tan2018mnasnet} and DPP-Net~\cite{dong2018dpp} propose formulations
that co-design the cell for both DNN accuracy and runtime. More specifically, 
DPP-Net proposes a progressive search for Pareto-optimal DNNs, where a probabilistic
model is obtained on smaller cell designs and new, more complex cells are being proposed 
by the probabilistic model as the search progresses. The DPP-Net NAS-based 
progressive model ranks candidate cell configurations based on a weighted ranking of 
both the  anticipated DL accuracy and the runtime of each configuration measured
on a hardware platform. Similarly, MnasNet considers a multi-objective term to be 
optimized, then accounts for both  runtime (on a Google Pixel phone platform) and accuracy
of candidate cell designs. 
\begin{figure}
  \centering
  \includegraphics[width=0.60\columnwidth]{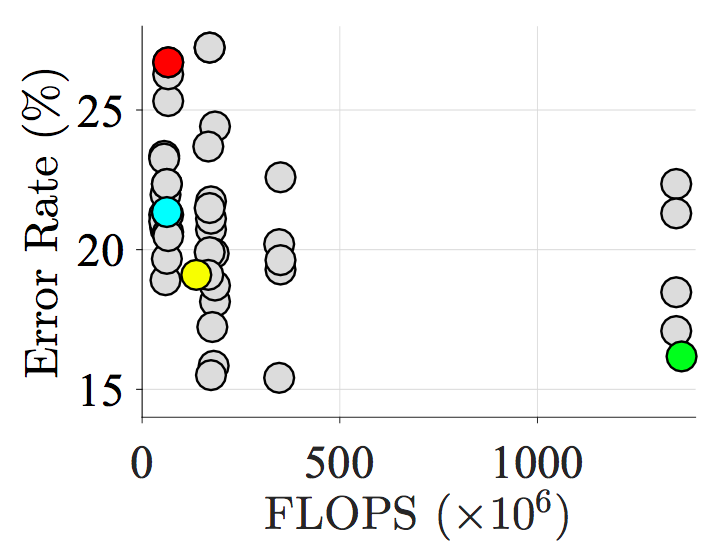}
  \vspace{-10pt}
  \caption{Pareto-optimal candidates (colored dots)
  identified by platform-aware NAS-based method~\cite{dong2018dpp}.}
  \vspace{-10pt}
  \label{fig:da-cheng}
\end{figure}
Figure \ref{fig:da-cheng} in \cite{dong2018dpp} shows the Pareto optimal DNN candidates with respect to error and FLOPS. The NAS-based method, \emph{i.e.}, DPP-Net, 
is able to identify models with a large number of parameters
and slow inference time. The DPP-Nets can achieve  better
trade-off among multiple objectives compared with state-of-the-art mobile CNNs and models designed using architecture search methods. Similarly, the MNAS method outperforms the state-of-the-art handcrafted designs, namely MobileNet~\cite{howard2017mobilenets}, by identifying designs closer to the Pareto front.

The ever-increasing interest of ML practitioners in NAS methods shows 
that these methodologies will provide the foundation for novel hardware-aware 
optimization techniques. Several open questions remain to be investigated, 
as we motivate further in the following section.

\section{Unexplored Research Directions}

\subsection{Hardware-aware modeling}

As motivated in the previous section, the development of predictive models
to capture the hardware performance of DL application has already played a critical 
role in the context of hyper-parameter optimization. To this end, we postulate that 
predictive models are poised to have significant impact towards enabling hardware-aware
ML. However, the following directions of research are crucial for applicability of these models in a co-design environment

\subsubsection{Hardware-aware models for DL with general topologies}

Current work has addressed building models largely for linear or pipelined neural network structures. However, as mentioned before, such predictive models should be suitably extended or developed anew for nonlinear 
DNN structures~\cite{cai2017neuralpower}. This is of critical importance especially in 
the context of NAS-based methods, where the cell design consists of several concatenating
operations that are executed in a nonlinear fashion with several non-sequential dependencies, 
unlike the case of traditional DNN designs. To this end, the development of hardware-aware predictive models for NAS-like frameworks is essential. 

\subsubsection{Cross-platform hardware-aware models}

Existing runtime and energy models currently consider specific types of platforms, with Nvidia GPUs being the hardware fabric in both the Paleo~\cite{qi17paleo} and the NeuralPower~\cite{cai2017neuralpower}
frameworks. It is therefore important to explore other platforms and segments of computing systems spanning the entire edge to server continuum. It becomes crucial to develop predictive models for capturing the runtime or energy consumption for other types of neural accelerators and platforms, 
such as reconfigurable architectures presented in~\cite{fowers2018brainwave}. Towards this direction, 
we postulate that cross-platform models or models that account for chip variability 
effects~\cite{cai2016exploring,stamoulis2016canwe,chen2017profit} can significantly help 
ML practitioners to transfer knowledge from one type of hardware platform to another and to choose the best model for that new platform. 

\subsection{Hardware-aware optimization}

There are several directions to advance state-of-the-art in the context of 
hardware-aware hyper-parameter optimization, especially given the 
recent interest in NAS-based formulations. 

\subsubsection{Multi-objective optimization} 
The design of DL models is a challenging task that requires different trade-offs 
beyond the currently investigated accuracy versus runtime/energy cases. For instance, 
while the use of a simpler DNN design can improve the overall runtime, it could
significantly degrade the utilization or throughput achieved given a fixed 
underlying hardware platform. Hence, we believe that several novel approaches
that focus on hardware-aware hyper-parameter optimization would be extending
current SMBO models to multiple design objectives. 

\subsubsection{Hardware-DL model co-optimization} We postulate that model-based hyper-parameter optimization 
approaches will allow innovation beyond the design of DL models, since 
the predictive models can be viewed also as a function of the underlying hardware 
platform. That is, Bayesian optimization- and NAS-based formulations can be extended
to cases where both the DL model and the hardware are co-designed. For instance, 
the optimal NAS-like cell can be identified while varying the hardware hyper-parameters of
a reconfigurable architectures presented, such as the number of processing 
elements~\cite{fowers2018brainwave}. Initial efforts focusing on the design 
of hardware accelerators based on 3D memory designs~\cite{gao2017tetris} already
exploit energy-based models~\cite{isscc_2016_chen_eyeriss} for design space exploration.

\section{Conclusion}

To conclude, tools and methodologies for hardware-aware machine learning have  increasingly attracted attention of both academic and industry researchers. In this paper, we have discussed recent work on modeling and optimization for various types of hardware platforms running DL algorithms and their impact on improving hardware-aware DL design. We point out several potential new directions in this area, such as cross-platform modeling and hardware-model co-optimization.

\begin{acks}
This research was supported in part by NSF CNS Grant No. 1564022 and by Pittsburgh Supercomputing Center via NSF CCR Grant No. 180004P.
\end{acks}

\bibliographystyle{ACM-Reference-Format}
\bibliography{dstam-iccad18}

\end{document}